\documentclass[12pt]{article}

\usepackage[utf8]{inputenc}
\usepackage[T1]{fontenc}
\usepackage{amsmath,amssymb,amsfonts}
\usepackage{graphicx}
\usepackage{booktabs}
\usepackage{hyperref}
\usepackage[margin=1in]{geometry}
\usepackage{natbib}
\usepackage{xcolor}

\hypersetup{colorlinks=true, linkcolor=blue, citecolor=blue, urlcolor=blue}

\newcommand{\bacc}{\text{BACc}}
\newcommand{\dhidden}{d_{\text{hidden}}}
\newcommand{\pp}{\,\text{pp}}
\newcommand{\xproj}{\mathbf{x}_{\text{proj}}}
\newcommand{\WIP}{\mathbf{W}_{\text{IP}}}
\newcommand{\WOP}{\mathbf{W}_{\text{OP}}}
\newcommand{\bIP}{\mathbf{b}_{\text{IP}}}
\newcommand{\bOP}{\mathbf{b}_{\text{OP}}}

\title{Analog Optical Inference on Million-Record Mortgage Data}

\author{%
  Sofia Berloff$^{1}$, Pavel Koptev$^{2}$, Konstantin Malkov$^{2}$\\[6pt]
  $^{1}$\textit{Department of Mathematics, University of York, York, UK}\\
  $^{2}$\textit{ai1Technologies, 27 Beach Rd Suite 5A Monmouth Beach, NJ 07750, USA}\\[4pt]
  \texttt{}
}

\date{}

\begin{document}
\maketitle

% ── ABSTRACT ──────────────────────────────────────────────────────────────
\begin{abstract}
Analog optical computers promise large efficiency gains for machine learning inference, yet no demonstration has moved beyond small-scale image benchmarks.
We benchmark the analog optical computer (AOC) digital twin on mortgage approval classification from 5.84~million U.S.\ HMDA records and separate three sources of accuracy loss.
On the original 19 features, the AOC reaches $94.6\%$ balanced accuracy with 5{,}126 parameters (1{,}024 optical), compared with $97.9\%$ for XGBoost; the $3.3$ percentage-point gap narrows by only $0.5\pp$ when the optical core is widened from 16 to 48 channels, suggesting an architectural rather than hardware limitation.
Restricting all models to a shared 127-bit binary encoding drops every model to $89.4$--$89.6\%$, with an encoding cost of ${\sim}8\pp$ for digital models and ${\sim}5\pp$ for the AOC.
Seven calibrated hardware non-idealities impose no measurable penalty.
The three resulting layers of limitation (encoding, architecture, hardware fidelity) locate where accuracy is lost and what to improve next.
\end{abstract}

\noindent\textbf{Keywords:} analog optical computing, deep equilibrium models, tabular classification, mortgage underwriting, hardware non-idealities, digital twin

% ── 1. INTRODUCTION ──────────────────────────────────────────────────────
\section{Introduction}
\label{sec:intro}

Machine learning inference accounts for roughly 90\% of computational energy in commercial deployments~\citep{kalinin2025aoc}, and the resulting demand for specialised hardware has renewed interest in analog optical computing.
Because matrix--vector multiplications can be carried out in the optical domain and nonlinear operations in analog electronics, the von Neumann bottleneck is avoided entirely~\citep{mcmahon2023optical,stroev2023analog,shen2017deep,xu2021optical,wetzstein2020inference}.
The analog optical computer (AOC)~\citep{kalinin2025aoc} demonstrated this on MNIST and Fashion-MNIST, where a calibrated digital twin agreed with the physical hardware on over 99\% of inputs at a projected efficiency of 500~TOPS\,W$^{-1}$.

Whether analog optical hardware can cope with the heterogeneous tabular datasets that dominate enterprise ML remains untested.
Gradient-boosted decision trees (GBDTs) remain the strongest single-model default on tabular data~\citep{grinsztajn2022tree,shwartzziv2022tabular,gorishniy2021revisiting}, though the gap to neural methods has narrowed~\citep{gorishniy2025tabm,erickson2025tabarena}.
We test the AOC on the U.S.\ Home Mortgage Disclosure Act (HMDA) dataset~\citep{hmda} (5.84~million records), evaluating it on both the original continuous features and a 127-bit Ising-encoded binarisation.
Our main contributions:
\begin{enumerate}
    \item On raw features, the AOC reaches $94.6\%$ balanced accuracy with 1{,}024 optical weights, $3.3\pp$ below XGBoost ($97.9\%$). Widening to 48 channels adds only $0.5\pp$, suggesting the gap is primarily architectural.
    \item On binarised features, four models converge to $89.4$--$89.6\%$ with Jaccard error overlap of $0.81$--$0.83$, saturating the encoding ceiling. On raw features, the same model pair has Jaccard $= 0.35$: binarisation collapses model diversity along with accuracy.
    \item Seven calibrated hardware non-idealities impose no penalty in either setting.
\end{enumerate}

% ── 2. BACKGROUND ────────────────────────────────────────────────────────
\section{Background: The Analog Optical Computer}
\label{sec:background}

The AOC hardware~\citep{kalinin2025aoc} implements a deep-equilibrium model (DEQ)~\citep{bai2019deq}.
The hidden state $\mathbf{s}_t \in \mathbb{R}^{\dhidden}$ evolves according to
\begin{equation}
    \mathbf{s}_{t+1} = \alpha\,\mathbf{s}_t + \beta\,\mathbf{W}\,\tanh(\mathbf{s}_t) + \mathbf{b} + \xproj,
    \label{eq:deq}
\end{equation}
where $\alpha = 0.5$ is a momentum coefficient fixed by the analog electronics (not a trainable hyperparameter), $\beta$ is a fixed gain, $\mathbf{W} \in \mathbb{R}^{\dhidden \times \dhidden}$ is the optical weight matrix, $\tanh$ is applied element-wise, $\mathbf{b} \in \mathbb{R}^{\dhidden}$ is a trained bias, and $\xproj = \WIP \mathbf{x} + \bIP$ is the input projection, with $\WIP \in \mathbb{R}^{\dhidden \times d_{\text{in}}}$ and $\bIP \in \mathbb{R}^{\dhidden}$.
The system iterates from $\mathbf{s}_0 = \mathbf{b} + \xproj$ until the state converges to a fixed point~$\mathbf{s}^*$ (relative L2 tolerance $10^{-3}$, typically ${\sim}9$ iterations on HMDA data).
The classification output is $\mathbf{y} = \WOP \mathbf{s}^* + \bOP$, where $\WOP \in \mathbb{R}^{2 \times \dhidden}$ and $\bOP \in \mathbb{R}^{2}$.

The input-projection ($\WIP$) and output-projection ($\WOP$) layers are standard digital linear transformations and accept any real-valued input $\mathbf{x} \in \mathbb{R}^{d_{\text{in}}}$.
Only the matrix--vector product $\mathbf{W}\,\tanh(\mathbf{s}_t)$ is performed optically, via microLED arrays and spatial light modulators.
The $\tanh$ nonlinearity is implemented in analog electronics.
On the physical hardware, $\mathbf{W}$ is quantised to 9-bit integers and each optical pass takes approximately 20\,ns.
For the four-block ensemble, each DEQ iteration requires four sequential passes on a single optical core (80\,ns per iteration); at ${\sim}9$ iterations the optical recurrence takes approximately 720\,ns~\citep{kalinin2025aoc}.

The AOC digital twin (AOC-DT) replicates the hardware as a differentiable PyTorch module.
The \textbf{AOCCell} variant incorporates seven calibrated non-idealities measured on the physical device:
(1)~tanh approximation error,
(2)~micro-LED input nonlinearity,
(3)~spatial light modulator (SLM) weight distortion,
(4)~transimpedance amplifier (TIA) gain variation,
(5)~photodiode crosstalk,
(6)~SLM darkness leakage, and
(7)~power normalisation.
The \textbf{SimpleCell} variant omits all seven, providing an idealised reference.

The current hardware has 16 optical channels ($16 \times 16$ weight matrix, 256 weights).
We use an ensemble architecture: \mbox{Ens-4x-AOC-$N$} runs four independent $N$-channel blocks via time-multiplexing; their $\dhidden$-dimensional outputs are concatenated into a $4N$-dimensional vector before the shared output-projection layer $\WOP$. We test $N = 16$ (1{,}024 optical weights) and $N = 48$ (9{,}216 optical weights).

% ── 3. METHODS ───────────────────────────────────────────────────────────
\section{Methods}
\label{sec:methods}

\subsection{Dataset and splitting}
\label{sec:dataset}

We use 5{,}843{,}255 public HMDA records~\citep{hmda} with binary disposition (originated/denied; 82\%/18\% class ratio).
The 19 raw features span loan characteristics (amount, term, interest rate, property value, combined LTV), borrower attributes (income, debt-to-income), loan type and purpose (6 binary indicators), and macroeconomic context (5 variables).

After stratified splitting (70/20/10) and downsampling the majority class to balance, we pool the balanced training and validation sets (1{,}906{,}780 samples) and re-split by \textit{group}: records sharing the same 127-bit binarised vector form a group assigned entirely to one partition, preventing data leakage.
This yields 1{,}524{,}335 training, 191{,}838 validation, and 190{,}607 test samples (Table~\ref{tab:splits}).
The same split is used for raw-feature experiments to ensure identical samples across both representations.

\begin{table}[t]
\centering
\caption{Group-split dataset. All splits are near class-balanced.}
\label{tab:splits}
\begin{tabular}{@{}lrrr@{}}
\toprule
Split & Samples & Unique groups & Class balance \\
\midrule
Train & 1{,}524{,}335 & ${\sim}$1{,}042{,}361 & 50.0\,/\,50.0\% \\
Validation & 191{,}838 & ${\sim}$130{,}295 & 50.3\,/\,49.7\% \\
Test & 190{,}607 & ${\sim}$130{,}295 & 49.8\,/\,50.2\% \\
\bottomrule
\end{tabular}
\end{table}

\subsection{Feature representations}
\label{sec:features}

\textbf{Raw features.}
For XGBoost, all 19 features are passed directly with ordinal-encoded categoricals.
For MLPs, categoricals are one-hot encoded (48 columns) and continuous features standardised, yielding 60 inputs.
For the AOC, the same 60 features are used with quantile-scaled continuous features (uniform on $[-1, +1]$) and Ising-encoded categoricals/binaries ($s = 2x - 1$, mapping $\{0,1\} \to \{-1,+1\}$).
The Ising centring is required by the $\tanh$-based DEQ: without it, sparse one-hot inputs (84\% zeros) produce IP-layer outputs with insufficient variance, trapping the model in the $\tanh$ linear regime.
In our experiments, omitting the centring (using $\{0,1\}$ instead of $\{-1,+1\}$) drops AOCCell accuracy to $71\%$, comparable to logistic regression.

\textbf{Binarised features.}
The 19 features are encoded into 127 Ising-valued columns: 6 continuous features are quantile-binned and one-hot encoded (66 bits), 7 categoricals are one-hot encoded with a catch-all bucket (55 bits), and 6 binaries pass through (6 bits).
All columns are mapped to $\{-1, +1\}$.

\subsection{Models}
\label{sec:models}

The AOC follows $\mathbf{x} \in \mathbb{R}^{d_{\text{in}}} \xrightarrow{\WIP} \mathbb{R}^{\dhidden} \xrightarrow{\text{DEQ}} \mathbf{s}^* \xrightarrow{\WOP} \mathbb{R}^{2}$, trained with Adam~\citep{kingma2015adam}, implicit differentiation~\citep{bai2019deq}, learning rate $3 \times 10^{-4}$, batch size 256, early stopping (patience 10, max 80 epochs), 3 seeds.
On raw features ($d_{\text{in}} = 60$): \mbox{Ens-4x-AOC-16} has 5{,}126 params (1{,}024 optical); \mbox{Ens-4x-AOC-48} has 21{,}510 params (9{,}216 optical).
On binarised features ($d_{\text{in}} = 127$): \mbox{Ens-4x-AOC-16} has 9{,}414 params (1{,}024 optical).

Classical baselines (3 seeds each):
XGBoost~\citep{chen2016xgboost} (1{,}000 estimators, max depth 5, lr 0.2, selected via randomised search with 3-fold CV on a 200K subsample);
MLP-small (input$\to$48$\to$48$\to$2, lr $5 \times 10^{-4}$);
MLP-large (input$\to$128$\to$128$\to$2, lr $10^{-3}$);
logistic regression (default regularisation).
MLP architectures and learning rates were chosen to bracket the AOC's parameter count but were not extensively tuned; the MLP baselines may therefore understate what an optimised neural network could achieve on this task.
More recent tabular deep learning methods (TabM~\citep{gorishniy2025tabm}, TabPFN~\citep{grinsztajn2025tabpfn}) were not tested, as the goal was to compare the AOC against standard baselines rather than to establish a state-of-the-art result.
All evaluated by balanced accuracy ($\bacc$, the arithmetic mean of per-class recall).

% ── 4. RESULTS ───────────────────────────────────────────────────────────
\section{Results}
\label{sec:results}

\subsection{Raw-feature performance}
\label{sec:rawresults}

\begin{table}[t]
\centering
\caption{Test balanced accuracy on 190{,}607 held-out mortgage applications, raw features (mean $\pm$ std over 3 seeds).
Optical parameters are the subset executed on analog hardware.}
\label{tab:raw}
\begin{tabular}{@{}llrrr@{}}
\toprule
Model & Features & Parameters & Optical & Test $\bacc$ (\%) \\
\midrule
XGBoost & 19 raw & ${\sim}$31{,}500 & --- & $97.91 \pm 0.00$ \\
MLP-large & 60 (one-hot) & 24{,}578 & --- & $97.46 \pm 0.04$ \\
MLP-small & 60 (one-hot) & 5{,}378 & --- & $97.21 \pm 0.17$ \\
\textbf{Ens-4x-AOC-48} & \textbf{60 (Ising)} & \textbf{21{,}510} & \textbf{9{,}216} & $\mathbf{95.14 \pm 0.46}$ \\
\textbf{Ens-4x-AOC-16} & \textbf{60 (Ising)} & \textbf{5{,}126} & \textbf{1{,}024} & $\mathbf{94.64 \pm 0.53}$ \\
Ens-4x-AOC-16 (SimpleCell) & 60 (Ising) & 5{,}122 & --- & $93.85 \pm 1.58$ \\
Logistic regression & 60 & 122 & --- & $70.04$ \\
\bottomrule
\end{tabular}
\end{table}

\begin{figure}[t]
    \centering
    \includegraphics[width=0.92\textwidth]{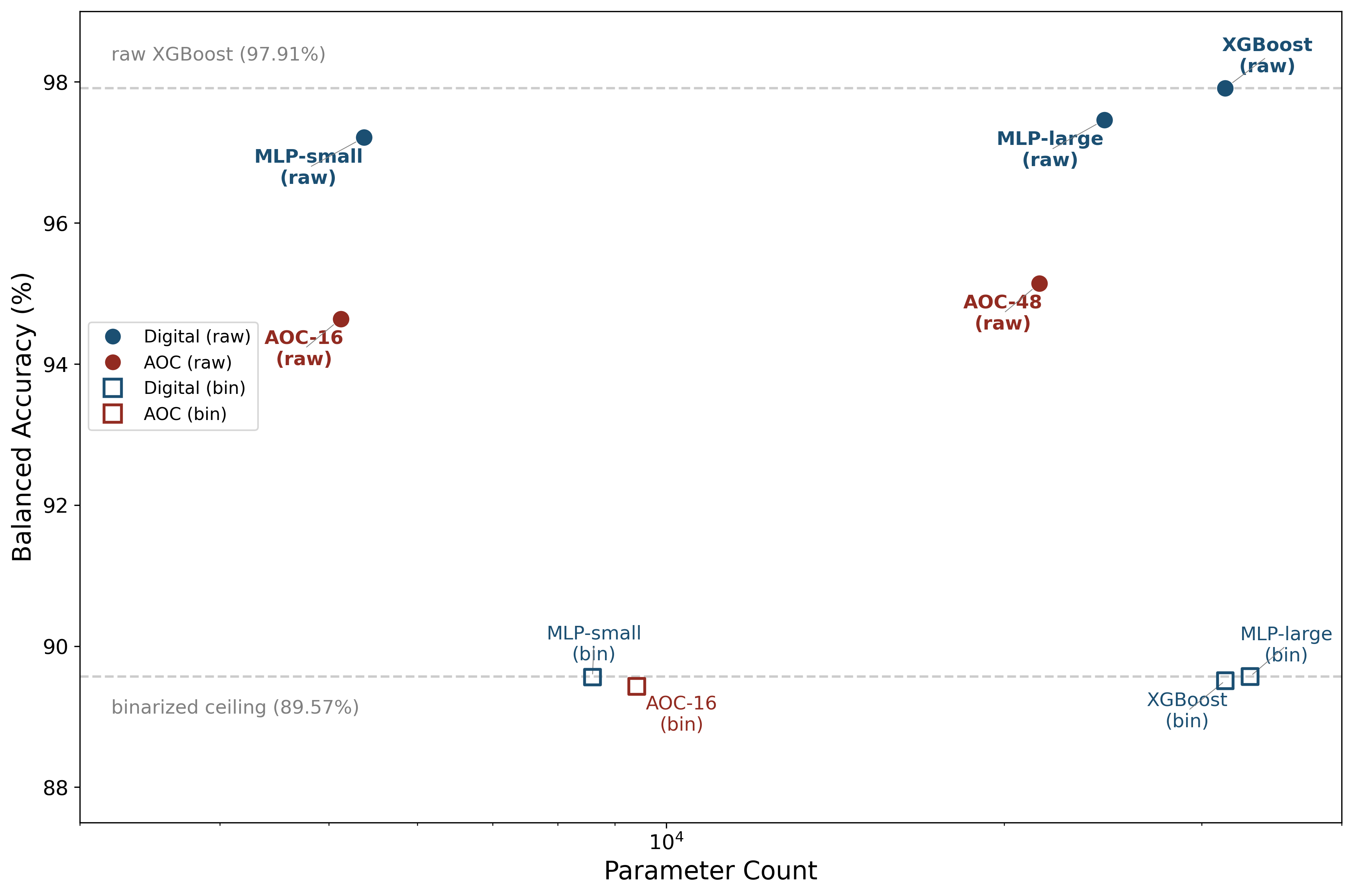}
    \caption{Balanced accuracy vs.\ parameter count.
    Filled circles: raw features; open squares: binarised.
    On raw features, AOC-16 and MLP-small have similar parameter counts (${\sim}5{,}200$) but a $2.6\pp$ gap.
    In the binarised setting, all nonlinear models collapse to $89.4$--$89.6\%$ regardless of capacity.}
    \label{fig:params}
\end{figure}

Table~\ref{tab:raw} summarises the main result.
XGBoost leads at $97.91\%$; the two MLPs reach $97.5\%$ and $97.2\%$.
The AOC (\mbox{Ens-4x-AOC-16}, 5{,}126 parameters, 1{,}024 optical) reaches $94.64 \pm 0.53\%$.
The wider \mbox{Ens-4x-AOC-48} (21{,}510 parameters, 9{,}216 optical) reaches $95.14\%$, only $+0.5\pp$ above the 16-channel model despite $4.2\times$ more parameters. This suggests the gap to XGBoost is primarily architectural rather than a channel-count limitation, though we note this conclusion rests on two channel widths; intermediate sizes (24, 32) would strengthen the evidence for saturation.
MLP-small (5{,}378 params, $97.2\%$) and AOC-16 (5{,}126 params, $94.6\%$) have nearly identical parameter counts but a $2.6\pp$ accuracy gap (Figure~\ref{fig:params}). The MLP uses two independent ReLU layers; the AOC reuses one $16 \times 16$ weight matrix across ${\sim}9$ iterations, gaining optical parallelism but giving up nonlinear capacity.
Part of this gap may also reflect differences in input preprocessing: the MLP receives standardised features with $\{0,1\}$ one-hot encoding, while the AOC receives quantile-scaled continuous features and $\{-1,+1\}$ Ising-encoded categoricals. We have not isolated the contribution of preprocessing from that of architecture.

\begin{figure}[t]
    \centering
    \includegraphics[width=0.78\textwidth]{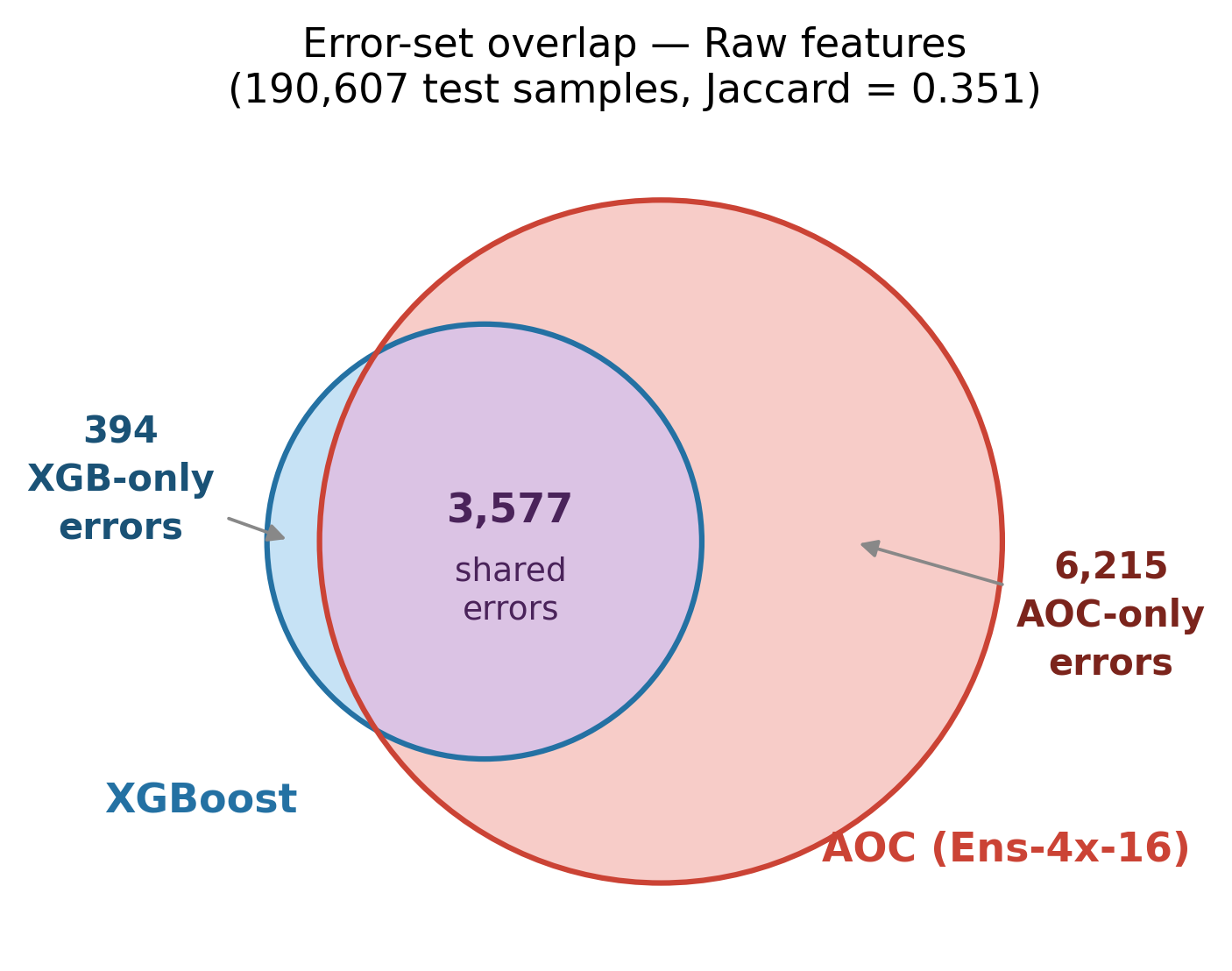}
    \caption{Error-set overlap on raw features (XGBoost seed~42 vs AOC seed~0).
    XGBoost makes 3{,}971 errors; the AOC makes 9{,}792; only 3{,}577 are shared (Jaccard $= 0.351$).
    On binarised features (Figure~\ref{fig:errors}) the same pair shares over $90\%$ of its errors (Jaccard $= 0.834$).
    Binarisation collapses model diversity along with accuracy.}
    \label{fig:errors_raw}
\end{figure}

The error-overlap analysis (Figure~\ref{fig:errors_raw}) shows that on raw features, XGBoost and the AOC share only $35\%$ of their errors (Jaccard $= 0.351$), compared with $83\%$ on binarised features.
The AOC makes $6{,}215$ errors that XGBoost gets right, while XGBoost makes only $394$ that the AOC gets right.
This asymmetry means the AOC's errors are nearly a superset of XGBoost's.
Despite the low Jaccard, a majority-vote ensemble of AOC and XGBoost does not improve over XGBoost alone: the $3.3\pp$ accuracy gap means the AOC's unique errors outnumber its unique corrections $16{:}1$, so naive voting cannot exploit the error complementarity.
The practical consequence of the low Jaccard is not better ensembles but a confirmation that the two architectures solve the problem differently, and that the AOC's failures are concentrated on cases that XGBoost finds easy.

\subsection{Effect of binarisation}
\label{sec:binresults}

\begin{table}[t]
\centering
\caption{Effect of binarisation (mean over 3 seeds). All models evaluated on the same 190{,}607 test samples.}
\label{tab:bincost}
\begin{tabular}{@{}lrrrr@{}}
\toprule
Model & Raw $\bacc$ (\%) & Binarised $\bacc$ (\%) & $\Delta$ (pp) \\
\midrule
XGBoost & 97.91 & 89.51 & $-8.40$ \\
MLP-large & 97.46 & 89.57 & $-7.89$ \\
MLP-small & 97.21 & 89.56 & $-7.65$ \\
Ens-4x-AOC-16 & 94.64 & 89.43 & $-5.21$ \\
Logistic regression & 70.04 & 76.54 & $+6.50$ \\
\bottomrule
\end{tabular}
\end{table}

\begin{figure}[t]
    \centering
    \includegraphics[width=0.78\textwidth]{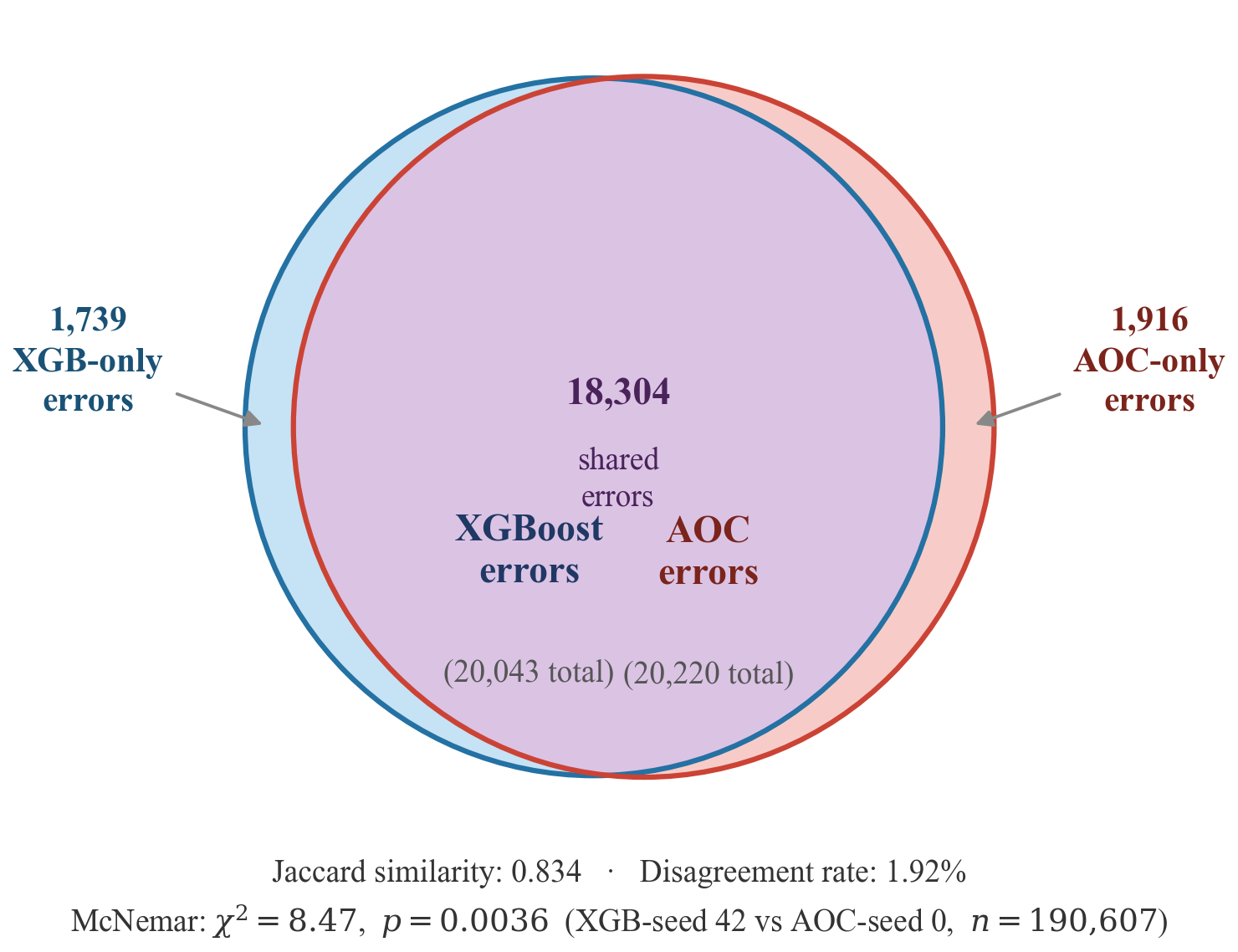}
    \caption{Error-set overlap on binarised features.
    Each model makes roughly 20{,}000 errors; 18{,}304 are shared (Jaccard $= 0.834$).
    The models disagree on just $1.92\%$ of the test set, having saturated the encoding ceiling.}
    \label{fig:errors}
\end{figure}

When all models are restricted to the 127-bit Ising encoding (Table~\ref{tab:bincost}), four nonlinear models converge to $89.4$--$89.6\%$, within $0.14\pp$ of one another.
The Jaccard similarity of AOC and XGBoost error sets jumps from $0.351$ (raw) to $0.834$ (binarised); McNemar's test~\citep{mcnemar1947} confirms statistically distinguishable predictions ($p < 0.05$) but the practical gap is at most $0.14\pp$ (Figure~\ref{fig:errors}).
Binarisation costs the digital baselines $7.7$--$8.4\pp$ and the AOC $5.2\pp$; the AOC loses less because its 16-channel compression already discards some information that the digital models exploit.
Logistic regression is the exception: binarisation helps by $+6.5\pp$, because one-hot binning creates linearly separable features.

\begin{figure}[t]
    \centering
    \includegraphics[width=0.88\textwidth]{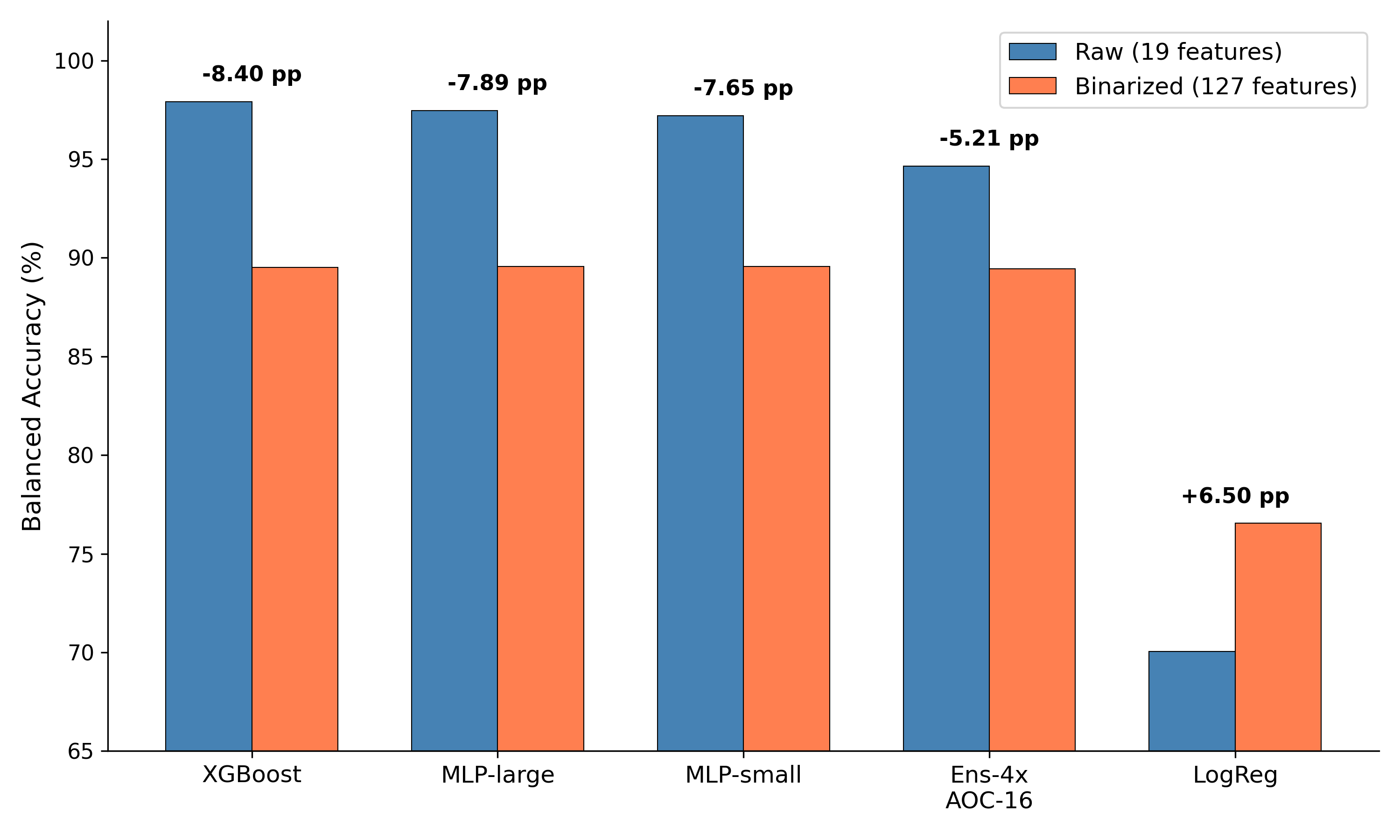}
    \caption{Raw vs.\ binarised balanced accuracy for all model families.
    Binarisation costs $7.7$--$8.4\pp$ for digital models and $5.2\pp$ for the AOC.
    Logistic regression gains $+6.5\pp$ because one-hot binning linearises the feature space.}
    \label{fig:rawvsbin}
\end{figure}

\subsection{Hardware non-idealities}
\label{sec:nonidealities}

On raw features, AOCCell ($94.64 \pm 0.53\%$) scores $0.8\pp$ higher than SimpleCell ($93.85 \pm 1.58\%$) with $3\times$ lower cross-seed variance, but with only 3 seeds per condition neither the mean difference nor the variance difference is statistically significant.
On binarised features, an 8-seed paired comparison at $\dhidden = 48$ finds no significant difference (paired $t$-test $p = 0.095$ on validation, $p = 0.950$ on test).
In both settings, the seven analog impairments do not hurt classification performance. The reversed direction is consistent with observations in \citep{kalinin2025analog}, where hardware non-idealities occasionally improved performance relative to the idealised model, possibly by acting as implicit regularisation; the lower cross-seed variance of AOCCell (0.53 vs 1.58 pp) supports this interpretation, though 3 seeds are insufficient to confirm it.

% ── 5. DISCUSSION ────────────────────────────────────────────────────────
\section{Discussion}
\label{sec:discussion}

\subsection{Three layers of limitation}
\label{sec:layers}

\begin{figure}[t]
    \centering
    \includegraphics[width=0.88\textwidth]{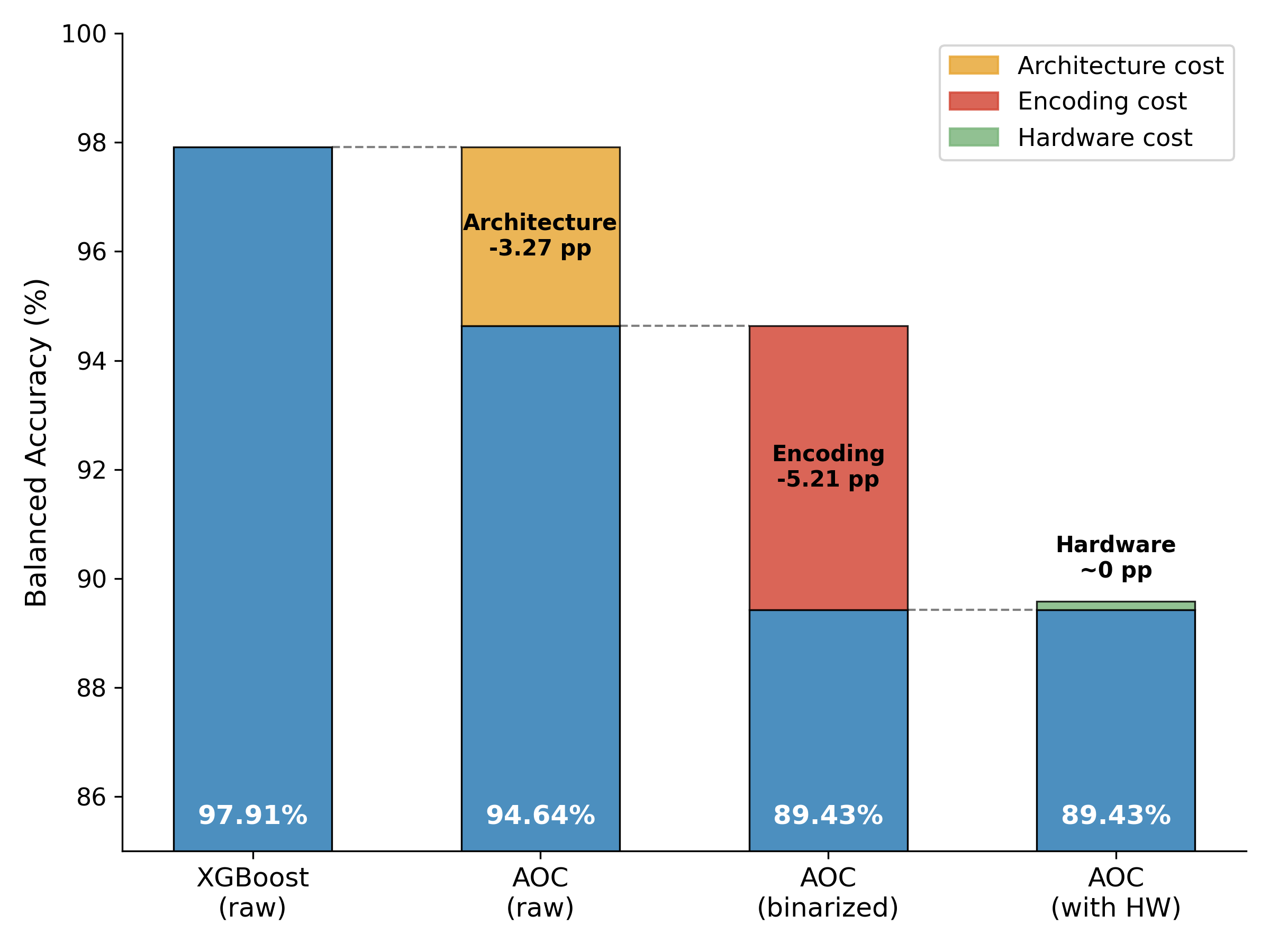}
    \caption{Decomposition of the accuracy gap between raw-feature XGBoost ($97.9\%$) and the binarised AOC ($89.4\%$).
    Architecture accounts for $3.3\pp$, encoding for $5.2\pp$, and hardware non-idealities for ${\sim}0\pp$.}
    \label{fig:layers}
\end{figure}

The comparison across two feature representations separates three sources of accuracy loss (Figure~\ref{fig:layers}).
\textbf{Encoding} (${\sim}8\pp$ for digital, ${\sim}5\pp$ for AOC): binarisation compresses the feature space until all nonlinear models hit the same ceiling (Section~\ref{sec:binresults}).
\textbf{Architecture} (${\sim}3\pp$): the DEQ's shared-weight recurrence gives optical parallelism but reduces nonlinear capacity. XGBoost can learn piecewise-constant boundaries on each feature independently, while the AOC's linear input projection compresses all features into a low-dimensional space. Matching GBDTs with neural methods on tabular data remains a persistent challenge~\citep{borisov2022deep,rubachev2025tabred}, though foundation-model approaches are beginning to close the gap~\citep{grinsztajn2025tabpfn}.
\textbf{Hardware non-idealities} (${\sim}0\pp$): seven calibrated impairments impose no penalty in either setting (Section~\ref{sec:nonidealities}).

The three costs are not fully independent: the AOC's lower encoding cost ($5.2$ vs $8.4\pp$ for XGBoost) partly reflects information already lost in the architectural compression. The decomposition is therefore approximate, but it identifies where the largest gaps are and which to address first.

\subsection{Implications}

Using raw features eliminates the encoding bottleneck entirely. The hardware-fidelity cost is zero. The remaining ${\sim}3\pp$ architectural gap is where future work should focus: deeper equilibrium networks, learned feature transformations, or hybrid AOC/digital architectures.
A single optical module supporting $N = 60$ variables would both eliminate the input-projection compression that drives this gap and remove the four-block time-multiplexing overhead, simultaneously improving accuracy and latency.

The binarised convergence result has a separate use. In domains where features are natively binary, such as molecular fingerprints in drug discovery, the Ising encoding is a natural representation, not a lossy compression, and the four-way convergence at $89.5\%$ shows that the AOC matches digital baselines in that regime.

\subsection{Inference latency}

The $3.3\pp$ accuracy gap to XGBoost comes with a projected latency advantage in the optical core.
On the 190{,}607-sample test set, XGBoost takes $3.7\,\mu$s per sample in batch mode and $190\,\mu$s for single-sample inference on a single-threaded Intel Core Ultra 7 265K CPU; on an NVIDIA RTX 5090 GPU, batch throughput improves to $0.14\,\mu$s per sample but single-sample latency rises to $1{,}731\,\mu$s owing to kernel launch overhead.
MLP-small takes $0.31\,\mu$s (batch) and $11.8\,\mu$s (single) on the same CPU; on the same GPU, single-sample latency rises to $39\, \mu$s, confirming the kernel-launch pattern.
On the physical hardware, each optical matrix--vector multiply takes approximately 20\,ns~\citep{kalinin2025aoc}.
The four-block ensemble is time-multiplexed on a single optical core, so each DEQ iteration requires four sequential passes (80\,ns per iteration).
At 9 iterations the optical recurrence completes in approximately 720\,ns, giving the optical core a $16\times$ advantage over MLP-small and $264\times$ over XGBoost for single-sample CPU inference---and over $2{,}400\times$ compared with GPU, where kernel launch overhead makes single-sample prediction slower, not faster.
Digital input and output projections (60$\to$64 and 64$\to$2 linear transforms) are executed on a separate digital controller; their latency depends on the controller architecture and is not included in this figure.

The per-pass time (20\,ns) is essentially independent of matrix dimension, while digital cost scales as $O(d_\text{hidden}^2)$.
At $d_\text{hidden} = 16$, the digital matrix--vector multiply is trivial and per-call overhead dominates; the optical advantage grows as the hardware scales to larger modules.
A single optical module with $N = 60$ (eliminating time-multiplexing) would reduce the optical recurrence to $9 \times 20\,\text{ns} = 180$\,ns: $66\times$ faster than MLP-small and over $1{,}000\times$ faster than XGBoost.
In mortgage underwriting, production models routinely incorporate hundreds of features beyond the 19 in the public HMDA dataset: credit bureau histories, property valuations, fraud indicators, macroeconomic time series and the required hidden dimension must scale accordingly: the current $60 \to 16$ compression already costs $3.3\pp$, and a $2{,}000$-feature model at the same ratio would need $d_\text{hidden} \approx 500$.
At the projected single-module scale of $N = 2{,}000$ variables ($4 \times 10^6$ weights) with 2\,GHz bandwidth~\citep{kalinin2025aoc}, 9 iterations would complete in ${\sim}4.5$\,ns, compared with ${\sim}2$\,ms for the equivalent digital computation on a single CPU core with a ratio exceeding $10^5$.
These projections assume that the optical path can be miniaturised as described in~\citet{kalinin2025aoc} and that the digital I/O projections are handled by colocated logic; neither has been demonstrated at this scale.

\subsection{Limitations}
\label{sec:limitations}

(i)~All experiments use the digital twin, not physical hardware. Agreement has been checked only on MNIST~\citep{kalinin2025aoc}. The seven non-ideality calibrations are device-level measurements (micro-LED response curves, SLM transfer functions) that do not depend on input content, so we expect them to transfer across tasks, but this has not been verified experimentally on HMDA data.
(ii)~AOCCell calibration exists only for $\dhidden \in \{16, 48\}$.
(iii)~HMDA is a single binary classification task.
(iv)~Physical latency and energy figures are projections from~\citet{kalinin2025aoc}; software inference times for XGBoost and MLP-small are measured on a single-threaded Intel Core Ultra~7 265K CPU and NVIDIA RTX~5090 GPU. The 720\,ns projection covers only the optical fixed-point recurrence; digital input and output projections add controller-dependent overhead.
(v)~The AOC requires Ising-centred inputs ($\{-1,+1\}$) for its $\tanh$-based architecture; alternative preprocessing may affect results.
(vi)~All results are reported over 3 random seeds; the AOC's cross-seed standard deviation ($0.5\pp$) implies a 95\% confidence interval of roughly $\pm 1.3\pp$ on the mean.

% ── 6. CONCLUSION ────────────────────────────────────────────────────────
\section{Conclusion}
\label{sec:conclusion}

The AOC digital twin reaches $94.6\%$ balanced accuracy on HMDA mortgage classification using raw features, with 1{,}024 optical weights.
Binarisation drops all models by $5$--$8\pp$ and collapses their error diversity (Jaccard from $0.83$ to $0.35$).
The three layers of limitation identified here (encoding, architecture, hardware fidelity) locate the accuracy loss at each stage: raw features remove the encoding penalty, the hardware non-idealities impose no cost, and the ${\sim}3\pp$ architectural gap to XGBoost is the open problem.

% ── DATA AVAILABILITY ────────────────────────────────────────────────────
\section*{Data Availability}
HMDA data are publicly available from the Consumer Financial Protection Bureau at \url{https://ffiec.cfpb.gov/data-publication/}.
The AOC digital twin code is available at \url{https://github.com/microsoft/aoc}.
Processed datasets and experimental scripts will be released upon publication.

% ── REFERENCES ───────────────────────────────────────────────────────────
\bibliographystyle{unsrtnat}

\end{document}